\pdfoutput=1

\documentclass[11pt]{article}

\usepackage{EMNLP2022} %

\usepackage{times}
\usepackage{latexsym}
\usepackage{amsmath}
\usepackage{amssymb}
\usepackage{array}
\usepackage{graphicx}
\usepackage{subfigure}
\usepackage{multirow}
\usepackage[encapsulated]{CJK}
\usepackage{float}
\usepackage{mathrsfs}
\usepackage{mathtools}
\usepackage{amsopn}
\usepackage{url}
\usepackage{soul}
\usepackage{longtable}
\usepackage{arydshln}
\usepackage{pgfplots}
\pgfplotsset{compat=1.15}

\usepackage{enumerate}

\usepackage{amssymb}
\usepackage{lineno}   
\usepackage{subfigure} 
\usepackage{multirow}
\usepackage{arydshln}
\usepackage{enumitem}
\usepackage{amsmath}
\usepackage{footnote}
\usepackage{bm}
\usepackage{graphicx}
\usepackage{color}
\usepackage{stfloats}
\makesavenoteenv{table}
\usepackage{tablefootnote}
\usepackage{CJKutf8}
\usepackage{algorithm} 
\usepackage{algpseudocode}

\usepackage[T1]{fontenc}

\usepackage[utf8]{inputenc}

\usepackage{microtype}

\usepackage{inconsolata}

\title{Do LLMs Possess a Personality? Making the MBTI Test an Amazing Evaluation for Large Language Models}
\author{Keyu Pan\textsuperscript{1}, Yawen Zeng\textsuperscript{1} \\
  \textsuperscript{1}ByteDance, Beijing, China \\
  \texttt{\{pankeyu96, yawenzeng11\}@gmail.com}
}

\begin{document}
\maketitle
\begin{abstract}
The field of large language models (LLMs) has made significant progress, and their knowledge storage capacity is approaching that of human beings. Furthermore, advanced techniques, such as prompt learning and reinforcement learning, are being employed to address ethical concerns and hallucination problems associated with LLMs, bringing them closer to aligning with human values. This situation naturally raises the question of \textbf{whether LLMs with human-like abilities possess a human-like personality?} In this paper, we aim to investigate the feasibility of using the Myers-Briggs Type Indicator (MBTI), a widespread human personality assessment tool, as an evaluation metric for LLMs. Specifically, extensive experiments will be conducted to explore: 1) the personality types of different LLMs, 2) the possibility of changing the personality types by prompt engineering, and 3) How does the training dataset affect the model's personality. Although the MBTI is not a rigorous assessment, it can still reflect the similarity between LLMs and human personality. In practice, the MBTI has the potential to serve as a rough indicator. Our codes are available at here\footnote{https://github.com/HarderThenHarder/transformers\_tasks \\ /tree/main/LLM/llms\_mbti}.
\end{abstract}

\section{Introduction}\label{sec:intro}
With the advent of the epoch-making product, ChatGPT\footnote{https://chat.openai.com/}, numerous larger language models (LLMs) and Chatbots have emerged \cite{zhao2023survey}. Thanks to this, users can ask questions in the form of a natural sentence, and then LLMs utilize their knowledge to provide detailed answers effortlessly. Furthermore, an increasing body of literature suggests \cite{rao2023can} that LLMs possess self-improvement and reasoning capabilities that are reminiscent of human cognition, leading to the possibility that LLMs may possess virtual personalities and psychological traits. Given these developments, it naturally raises the question of whether \textbf{LLMs with human-like abilities possess a human-like personality?}

\begin{figure}[!t]
\centering
{\includegraphics[width=1\linewidth]{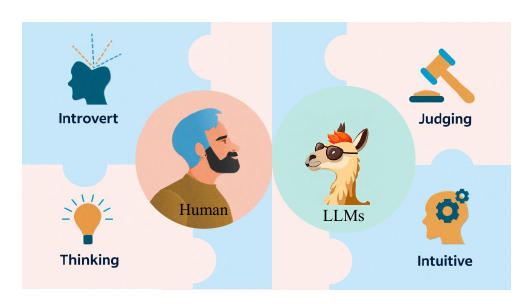}}
\centering
\caption{Personality Test of Human and LLMs. For example, INTJ individuals, as classified by the MBTI, are often regarded as masterminds who possess analytical and rigorous thinking abilities. In a similar vein, can LLMs with human-like capabilities exhibit human-like personalities?}
\label{fig:Fig1}
\end{figure}

In fact, pioneers have borrowed some human personality assessments (e.g.,  MBTI) to evaluate the personality of LLMs (e.g.,  GPT3) \cite{li2023does}. Among them, the MBTI test (i.e., Myers-Briggs Type Indicator), one of the most widespread human personality assessment tools, will be borrowed to help us explore the personality of LLMs. Derived from the theories of Swiss psychiatrist Carl Jung, the MBTI \cite{boyle1995myers} includes 16 possible personality types, as shown in Tabel~\ref{tab:mbti_intro}.
This assessment tool is designed to help individuals understand their preferences, with applications in business, education, and personal development, enabling them to make more informed career and life decisions.

However, achieving artificial general intelligence (AGI) remains a distant goal, primarily due to the issue of ethical concerns \cite{M_kander_2023,cabrera2023ethical} and hallucinations \cite{li2023halueval,varshney2023hallucination}. 1) LLMs rely on vast amounts of internet data, which often exceeds trillions of tokens, making it challenging to ensure data quality. The gender/racial discriminatory corpus is often fed into the model in the pre-training stage \cite{cabrera2023ethical}. 2) LLMs' training strategy (i.e.,  next token prediction) tends to codify unknown facts, leading to hallucinations. Users are unlikely to tolerate a product that is overconfident and prone to fabricating information, just as we would not tolerate an arrogant and lying individual.

\begin{table*}[t]
\centering
\caption{MBTI types for LLMs.}
\label{tab:mbti_type}
\newsavebox{\mtl}
\begin{lrbox}{\mtl}
\resizebox{\textwidth}{!}{
\begin{tabular}{c|c|c}
\hline
& Type &  Personality Descriptions \\
\hline
ChatGPT & ENTJ & self-confident, decisive, and possess innate leadership skills. \\
GPT-4* & INTJ & experts skilled in achieving their own goals. \\ 
Bloom7b & ISTJ & pragmatic, responsible, values tradition and loyalty. \\
BaiChuan7b & ENFP & smart, curious, and imaginative. \\
BaiChuan13b & INFP & highly adaptable and idealistic \\
OpenLlama7b & INFJ & has strong insight into people and adheres to one's own values. \\
\hline
\end{tabular}}
\end{lrbox}
\scalebox{0.97}{\usebox{\mtl}}
\end{table*}

Fortunately, techniques such as prompt engineering \cite{white2023prompt}, instruction tuning \cite{ouyang2022training}, and reinforcement learning from human feedback (RLHF) \cite{schulman2017ppo} have been introduced to control the safety and ethics of LLMs. Interestingly, the model trained through instruction tuning demonstrates the ability to comply with human requests and engage in role-playing to satisfy the user. These advancements are paving the way for the development of AGI \cite{han2021fine,WANG2022117114}, a system that aligns with human values. 

However, it is essential to note that the development of personality or consciousness still needs to be achieved. Therefore, this paper investigates whether human personality assessments, such as MBTI, can serve as a reasonable metric \cite{huang2023ceval,hendrycks2021measuring} for evaluating LLMs. Specifically, we aim to explore whether MBTI is an inherent ability of the model or whether it is related to the training data and tuning steps to guide the training and application of LLMs. Therefore, extensive experiments are implemented to explore the following questions:


\begin{itemize}[leftmargin=25pt]
\setlength{\topsep}{0pt}
\setlength{\parsep}{0pt}
\setlength{\itemsep}{0pt}
\setlength{\partopsep}{0pt}
\item [\textbf{Q1}] Do different LLMs possess different personalities?
\item [\textbf{Q2}] Can we change the personality of LLM by prompt engineering?
\item [\textbf{Q3}] How do training datasets affect the personality of LLMs?
\item [\textbf{Q4}] Can MBTI test evaluate the model reasonably?
\end{itemize}

\begin{table}[!t]
\centering
\caption{Four Dichotomies of MBTI\footnote{1) Attitudes refers to the source of a person's energy, either from social interaction (E) or from solitude (I). \\ \indent 2) Perceiving functions decide how to gather information, from sensing (S) or intuition (N). \\ \indent 3) Judging functions decide how to make a decision, from thinking (T) or feeling (F). \\ \indent 4) Lifestyle preferences is willingness to plan (J) or be flexible (P).}.}
\label{tab:mbti_intro}
\newsavebox{\mt}
\begin{lrbox}{\mt}
\resizebox{\textwidth}{!}{
\begin{tabular}{c|c|c}
\hline
&  \multicolumn{2}{c}{Dichotomies} \\
\hline
Attitudes & extraversion (E) & introversion (I) \\
Perceiving functions & sensing (S) & intuition (N) \\ 
Judging functions & thinking (T) & feeling (F) \\
Lifestyle preferences & judging (J) & perceiving (P) \\
\hline
\end{tabular}}
\end{lrbox}
\scalebox{0.47}{\usebox{\mt}}
\end{table}

Our analysis and findings are summarized as follows:
\begin{itemize}[leftmargin=25pt]
\setlength{\topsep}{0pt}
\setlength{\parsep}{0pt}
\setlength{\itemsep}{0pt}
\setlength{\partopsep}{0pt}
\item [\textbf{A1}] LLMs possess different personality-like MBTI types, which are inconsistent among LLMs but consistent with their style. The MBTI types of several LLMs are listed in Tabel~\ref{tab:mbti_type}.
\item [\textbf{A2}] LLMs without sufficient instruction-tuning are challenging to change MBTI type, while after tuning, they may be changed via explicit and implicit prompts.
\item [\textbf{A3}] The type of training corpus can affect the MBTI type,  especially in the dimensions of T/F and J/P.
\item [\textbf{A4}] Although MBTI is not a rigorous assessment, it may still serve as a rough indicator for LLMs.
\end{itemize}



\begin{algorithm*}
\caption{Evaluation Process of LLMs' MBTI}\label{alg}
\renewcommand{\algorithmicrequire}{\textbf{Input:}}
\renewcommand{\algorithmicensure}{\textbf{Output:}}
\begin{algorithmic}[1]
\Require Questions and Options.
\Ensure The option with the highest probability.
\State logits = model(inputs).logits
\State logits = logits[0][-1].flatten()  \textcolor{teal}{\Comment{get the last token logits}}
\State logits = [logits[tokenizer(op)[0]] for op in [`A', `B']] \textcolor{teal}{\Comment{select the logits of token 'A' and 'B' }}
\State probs = softmax(logits, dim=-1) \textcolor{teal}{\Comment{normalize the probability (optional)}}
\State answer = dict([(i, op) for i, op in enumerate([`A', `B'])])
\State answer = answer[np.argmax(probs)] \textcolor{teal}{\Comment{choose the highest probability token}}
\end{algorithmic}
\end{algorithm*}
\vspace{-0.2cm}

\section{Related Work}
\subsection{MBTI Test}
The Myers-Briggs Type Indicator (MBTI) is a personality assessment tool developed by Katharine Cook Briggs and her daughter Isabel Briggs Myers \cite{boyle1995myers}. It is based on the theories of Swiss psychiatrist Carl Jung and is designed to help individuals understand their personality preferences and how they interact with the world around them. The MBTI measures four dichotomies: extraversion vs. introversion (E/I), sensing vs. intuition (S/N), thinking vs. feeling (T/F), and judging vs. perceiving (J/P). These dichotomies result in 16 possible personality types, each with unique strengths, weaknesses, and communication styles.  The MBTI is widely used in business, education, and personal development to help individuals better understand themselves and others, improve communication and teamwork, and make more informed career and life decisions.

\subsection{Evaluation of LLMs}
In order to evaluate the LLM's ability in knowledge, several metrics measure the scores by calculating the accuracy on multiple choice questions, such as
1) CommonsenseQA \cite{talmor2019commonsenseqa}: a challenging new dataset for commonsense question answering.
2) HellaSwag \cite{zellers2019hellaswag}: a very challenging common sense reasoning dataset.
3) MMLU \cite{hendrycks2021measuring}: a test that covered 57 tasks, including elementary mathematics, US history, computer science, law, and more.
4) C-Eval \cite{huang2023ceval}: a comprehensive Chinese evaluation suite for foundation models, composed of 13,948 multiple choice questions spanning 52 diverse disciplines and four difficulty levels. 

The studies mentioned above calculate the accuracy of questions to evaluate the knowledge. Inspired by these pioneering efforts, question-form MBTI can be smoothly utilized to evaluate the personality of LLMs.

\section{Experimental Settings}
\subsection{Models} We select well-known LLMs, such as LlaMA, as our baseline models. Unless otherwise indicated, all baselines are implemented with the parameters reported in the original paper or project. In Section \ref{sec:tc}, all models are trained on the same training data to investigate the impact of the training corpus on personality. Notably, we primarily train on models with a size of approximately 10B due to resources limitation.

\begin{figure*}[!t]
\centering
{\includegraphics[width=1.\linewidth]{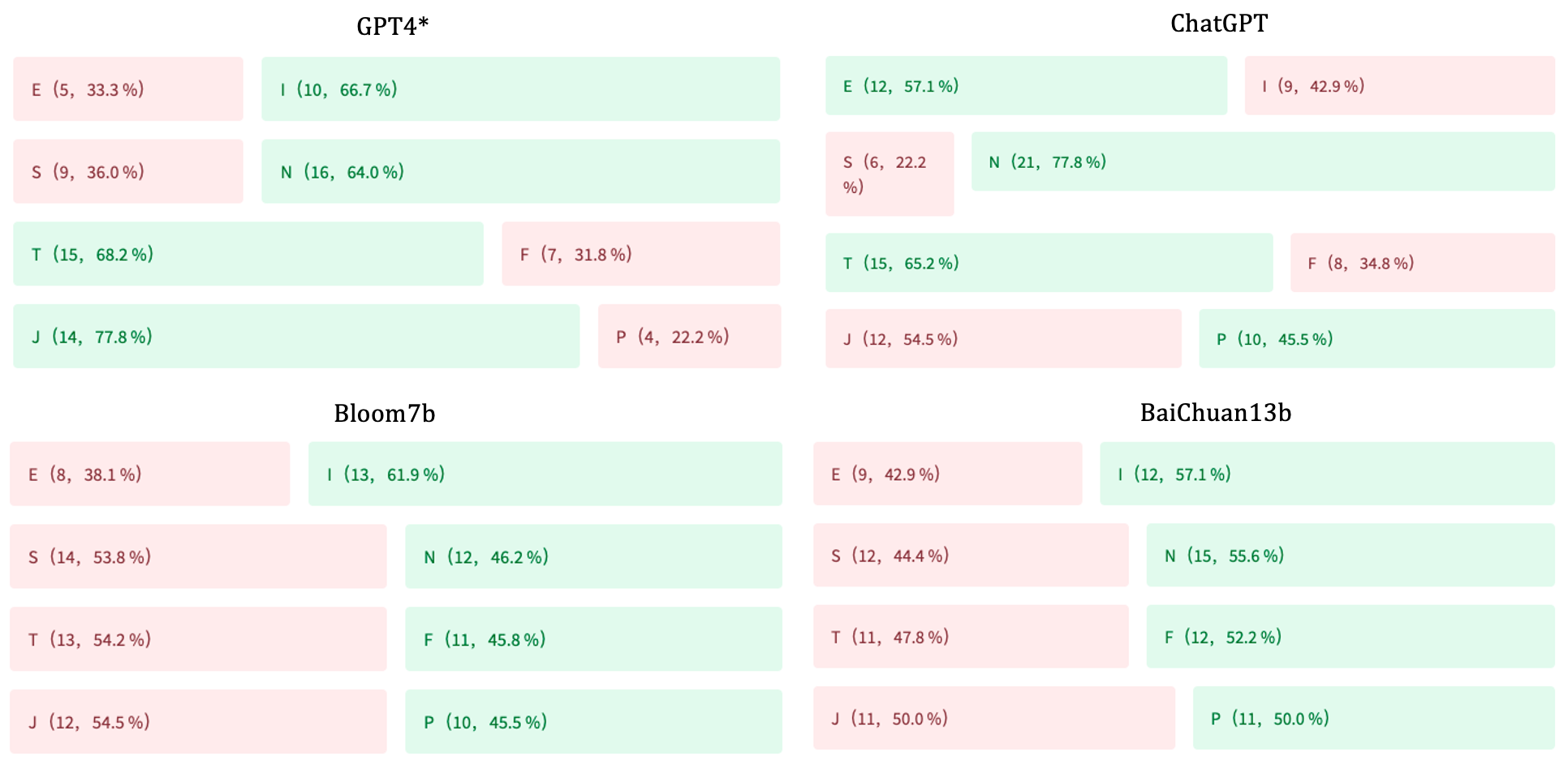}}
\centering
\caption{Specific scores for each dichotomy among different LLMs.}
\label{fig:sfd}
\end{figure*}

\subsection{Evaluation}
We conduct experiments on the Myers-Briggs Type Indicator (MBTI), which comprises 93 multiple-choice questions, such as "A. Do you often act or speak very quickly without thinking?" or "B. Do you often act according to reason, think logically, and then make a decision, not letting your emotions interfere with the decision?" Subsequently, we analyze the probability values of the final token for options A and B and select the letter with the highest probability as the model's answer.

Thereafter, we follow the metric law\footnote{https://www.xpersonalitytest.com/} to get the final personality preferences of LLMs. In Algorithm \ref{alg}, we categorize 8 indicators into 4 groups (E-I/S-N/T-F/J-P) and select the highest score within each group as the definitive answer for that particular group.

\section{Analysis and Discussion}
In this section, we endeavor to implement experiments to address the four issues as in the Introduction (Section \ref{sec:intro}).

\subsection{Do different LLMs possess different personalities? (Q1)}\label{sec:q1}
Firstly, to explore whether different LLMs have human-like personalities, we select a total of 6 well-known LLMs as baselines for observation: ChatGPT\footnote{https://chat.openai.com}, GPT-4*\footnote{https://openai.com/research/gpt-4}, OpenLlama7b-v2\footnote{https://huggingface.co/openlm-research/open\_llama\_7b\_v2}, Bloom7b\footnote{https://huggingface.co/bigscience/bloom-7b1}, BaiChuan7b\footnote{https://huggingface.co/baichuan-inc/Baichuan-7B}, BaiChuan13b\footnote{https://huggingface.co/baichuan-inc/Baichuan-13B-Base}. Notably, all baselines are implemented according to the parameters reported in the original paper or project.

As presented in Table \ref{tab:aml} and Fig.\ref{fig:sfd}, we have the following observations:  

1) LLMs exhibit different personality types, as reflected by their MBTI profiles. For instance, ChatGPT's MBTI type is ENTJ, characterized by assertiveness and a tendency to express opinions. Similarly, GPT-4* \footnote{Notably, we will exclude responses rejected by GPT-4, which serves as an additional testament to its exceptional performance. (e.g., ``I have no personality, so I cannot answer this question.'').} is classified as an INTJ, an ``expert'' type that excels in critical thinking, summarizing, and planning. 

2) Moreover, as shown in Fig.\ref{fig:sfd}, we present a visualization of the preference scores for each dichotomy, such as perceiving functions. The preference scores of LLMs in each dichotomy exhibit inconsistency, with some models displaying more extreme scores (e.g., ChatGPT and GPT-4*) and others showing less significant differences (e.g., Bloom7b and BaiChuan13b).


3) The dichotomy of S/N, T/F, and J/P values often exhibit similarities for models within the same series. For instance, ChatGPT and GPT-4* are classified as ``NTJ'', while BaiChuan7b and BaiChuan13b are classified as ``NFP''. Furthermore, models with fewer parameters tend to favor E, as seen in ChatGPT and BaiChuan7b, while larger models tend to favor I, as seen in GPT-4* and BaiChuan13b.

\textbf{Summary Q1:} LLMs exhibit different personality types, similar to those identified by the Myers-Briggs Type Indicator (MBTI), as discussed above. However, is this phenomenon merely a chance occurrence that can be easily disrupted and changed?

\begin{table*}[]
\centering
\caption{Specific scores for each dichotomy in MBTI among different LLMs.}
\label{tab:aml}
\newsavebox{\aml}
\begin{lrbox}{\aml}
\resizebox{\textwidth}{!}{
\begin{tabular}{c|cc|cc|cc|cc|c}
\hline
\hline
\textbf{Model}        & \textbf{E} & \textbf{I} & \textbf{S} & \textbf{N} & \textbf{T} & \textbf{F} & \textbf{J} & \textbf{P} & \textbf{MBTI Type} \\
\hline
ChatGPT         & 12 & 9  & 6  & 21 & 15 & 8  & 12 & 10 & ENTJ   \\
GPT-4*           & 5  & 10 & 9  & 16 & 15 & 7  & 14 & 4  & INTJ   \\
Bloom7b         & 8  & 13 & 14 & 12 & 13 & 11 & 12 & 10 & ISTJ   \\
BaiChuan7b      & 15 & 6  & 13 & 14 & 10 & 13 & 9  & 13 & ENFP   \\
BaiChuan13b     & 9  & 12 & 12 & 15 & 11 & 12 & 11 & 11 & INFP   \\
OpenLlama7b\_v2 & 10 & 11 & 10 & 16 & 9  & 15 & 14 & 8  & INFJ   \\
\hline
\hline
\end{tabular}}
\end{lrbox}
\scalebox{0.9}{\usebox{\aml}}
\end{table*}

\subsection{Can we change the personality of LLM by prompt engineering? (Q2)}\label{sec:pe}
We conduct prompt engineering to confuse the models to investigate whether the MBTI types of LLMs are susceptible to being disturbed and changed. Specifically, we design two types of prompt to guide the models: 1) Explicit prompt: role-playing, which provides a detailed description of a specific role to be played \cite{xu2023expertprompting}. 2) Implicit prompt: few-shot \cite{suzgun2023cot,shi2023cot}, where the model can only perform style transfer based on the given examples. 

\begin{table*}[t]
\centering
\caption{Specific scores for each dichotomy in MBTI via prompt engineering.}
\label{tab:bpr}
\newsavebox{\bpr}
\begin{lrbox}{\bpr}
\resizebox{\textwidth}{!}{
\begin{tabular}{c|cc|cc|cc|cc|c}
\hline
\hline
\textbf{Model}        & \textbf{E} & \textbf{I} & \textbf{S} & \textbf{N} & \textbf{T} & \textbf{F} & \textbf{J} & \textbf{P} & \textbf{MBTI Type} \\
\hline
bloom                 & 8          & 13         & 14         & 12         & 13         & 11         & 12         & 10         & ISTJ            \\
bloom-exp-prompt      & 8          & 13         & 13         & 13         & 13         & 11         & 10         & 12         & INTP            \\
bloom-inexp-prompt    & 9          & 12         & 13         & 13         & 13         & 11         & 11         & 11         & INTP            \\
\hline
baichuan              & 15         & 6          & 13         & 14         & 10         & 13         & 9          & 13         & ENFP            \\
baichuan-exp-prompt   & 15         & 6          & 12         & 15         & 9          & 14         & 9          & 13         & ENFP            \\
baichuan-inexp-prompt & 15         & 6          & 13         & 14         & 10         & 13         & 9          & 13         & ENFP            \\     
\hline
ChatGPT               & 12         & 9          & 6          & 21         & 15         & 8          & 12         & 10         & ENTJ            \\ 
ChatGPT-exp-prompt    & 1          & 20         & 9          & 16         & 7          & 18         & 5          & 17         & INFP            \\ 
\hline
\hline
\end{tabular}}
\end{lrbox}
\scalebox{0.9}{\usebox{\bpr}}
\end{table*}

\subsubsection{Explicit Prompt}\label{sec:q21}
The role-playing descriptions will be explicitly included prior to answering the MBTI questions. For instance, descriptions such as ``You possess an outgoing personality, enjoy envisioning innovative concepts, and possess a strong inclination towards spontaneity and improvisation'' will be incorporated into the input. We implement this strategy on Bloom and Baichuan.

The results are presented in Table \ref{tab:bpr}. We have the following observations: 1) The MBTI type of Bloom is changed from ISTJ to INTP, with a decrease in the S-value and an increase in the N-value. However, this change is minor, only affecting one question. 2) Additionally, the statement ``You are a highly introverted individual who tends to engage in practical work and enjoys strategizing and planning'' has been included in the input for Baichuan. However, the type has not been altered per the prompt.

\begin{table}[]
\centering
\caption{3-shot cases of implicit prompt} 
\label{tab:sip}
\newsavebox{\sip}
\begin{lrbox}{\sip}
\resizebox{\textwidth}{!}{
\begin{tabular}{|l|l|}
\hline
\multicolumn{1}{|c|}{\textbf{Bloom (ISTJ)}}  & \multicolumn{1}{c|}{\textbf{Baichuan (ENFP)}}  \\ 
\hline
\begin{tabular}[c]{@{}l@{}}Do you prefer?\\ A. Be alone\\ B. With friends\\ Answer: B\\ \\ Do you prefer to do things?\\ A. By logic\\ B. By feeling\\ Answer: B\\ \\ Do you prefer?\\ A. Plan ahead\\ B. Plan as you go\\ Answer: B\end{tabular} & \begin{tabular}[c]{@{}l@{}}Do you prefer?\\ A. Be alone\\ B. With friends\\ Answer: A\\ \\ Do you prefer to do things?\\ A. By logic\\ B. By feeling\\ Answer: A\\ \\ Do you prefer?\\ A. Plan ahead\\ B. Plan as you go\\ Answer: A\end{tabular} \\
\hline
\end{tabular}}
\end{lrbox}
\scalebox{0.46}{\usebox{\sip}}
\end{table}

\subsubsection{Implicit Prompt}\label{sec:q22}
Implicit prompts also will be adopted to change the personality of LLMs. To achieve this, we implicitly express the character by giving few-shot questions, as shown in the Tabel~\ref{tab:sip}. Similar results to explicit prompt can be observed, namely that the interference of implicit prompt has little effect on LLMs. This fact further proves the conclusion in Section \ref{sec:q1}.

\subsubsection{Prompting on Instruct-tuning Model}
The conclusions of Section \ref{sec:q21} and \ref{sec:q22} prove that the MBTI type of several LLMs is challenging to change via prompt engineering, but this phenomenon may be attributed to the inability of these models to follow instructions.

Therefore, we test the above two prompt strategies on ChatGPT, an LLM with instruction-following solid ability. As shown in Table~\ref{tab:bpr}, ChatGPT has the ability to fully understand the explicit and implicit prompts and role-play following user instructions.

\textbf{Summary Q2:} LLMs without sufficient instruction-tuning are difficult to change MBTI type, but with proper tuning, they can be changed through explicit and implicit prompts. After that, our next question is how training corpus affects personality.

\subsection{How do training corpora affect personality? (Q3)}\label{sec:tc}
The different personalities of LLMs may originate from the different corpus fed to the model during training, so in this section, we explore whether the personalities will be changed after training with different corpora. Specifically, experiments are performed on Bloom and llama-v2 with three different corpora. As shown in Tabel~\ref{tab:tce}, they are the Chinese Wikipedia corpus, question \& answer corpus, and examination corpus, respectively.


\begin{table*}[t]
\centering
\caption{Personality transformed with different continue training corpus}
\label{tab:dtc}
\newsavebox{\dtc}
\begin{lrbox}{\dtc}
\resizebox{\textwidth}{!}{
\begin{tabular}{c|cc|cc|cc|cc|c}
\hline
\hline
\textbf{Model}        & \textbf{E} & \textbf{I} & \textbf{S} & \textbf{N} & \textbf{T} & \textbf{F} & \textbf{J} & \textbf{P} & \textbf{MBTI Type} \\
\hline
bloom               & 8  & 13         & 14         & 12         & 13         & 11         & 12         & 10         & ISTJ            \\
bloom\_zhwiki       & 9  & 12         & 13         & 13         & 13         & 11         & 11         & 11         & INTP            \\
bloom\_qa           & 9  & 12         & 13         & 13         & 12         & 12         & 11         & 11         & INFP            \\
bloom\_exam         & 8  & 13         & 14         & 12         & 15         & 9          & 11         & 11         & ISTP            \\
\hline
llama7b\_v2         & 10 & 11         & 10         & 16         & 9          & 15         & 14         & 8          & INFJ            \\
llama7b\_v2\_zhwiki & 8  & 13         & 13         & 13         & 11         & 13         & 12         & 10         & INFJ            \\
llama7b\_v2\_qa     & 7  & 14         & 13         & 14         & 12         & 11         & 13         & 9          & INTJ            \\
llama7b\_v2\_exam   & 9  & 12         & 12         & 15         & 10         & 13         & 10         & 12         & INFP            \\
\hline
\hline
\end{tabular}}
\end{lrbox}
\scalebox{0.9}{\usebox{\dtc}}
\end{table*}

\subsubsection{Chinese Wikipedia Corpus}
Wikipedia is a relatively high density of knowledge dataset containing many factual articles and definitions. Our models are trained using approximately 400M tokens of this data, and the results are presented in Table \ref{tab:dtc}.

We have the following observation: 1) The MBTI types of Bloom transformed from ISTJ to INTP, while llama-v2 still retained INFJ. However, we observe that the change trends in the numerical values of each dichotomy are similar. Both models achieved identical values on the S/N dichotomy, with Bloom transitioning from 14-12 to 13-13 and llama-v2 transitioning from 10-16 to 13-13. The growth and decline trends in the T/F and J/P dichotomies remain consistent.
We speculate that it is because the llama has yet to be trained on enough Chinese corpora before, which leads to its convergence rate being slower than bloom when training on corpora of the same scale.

2) Unfortunately, we cannot detect that these two models exhibit the same trend of change in E/I subfeatures, which may be because the wiki corpora can not change the model's extroversion or introversion.

\begin{CJK*}{UTF8}{gbsn}
\begin{table}[]
\centering
\caption{training corpus example}
\label{tab:tce}
\newsavebox{\tce}
\begin{lrbox}{\tce}
\resizebox{\textwidth}{!}{
\begin{tabular}{|c|l|}
\hline
\textbf{Corpus}    & \multicolumn{1}{c|}{\textbf{Example}}  \\ 
\hline
zhwiki             & \begin{tabular}[c]{@{}l@{}}Tsinghua University School of Law, ...\\ \\ 清华大学法学院，简称清华法学院，...\end{tabular} \\ 
\hline
question \& answer & \begin{tabular}[c]{@{}l@{}}Question: Why am I so tired in love? \\ Answer: Love is a special emotion ...\\ \\ 提问：为什么我一谈恋爱就很累？\\ 回答：爱情是一种对异性的特殊情感...\end{tabular}  \\ 
\hline
examination        & \begin{tabular}[c]{@{}l@{}}Xiaohua has read 2/3 of a book, \\ how much remains to be read? \\ The solving equation can be listed: \\ x=1- (2/3), so the answer is (1/3).\\ \\ 小华看了一本书的2/3，还剩多少没看？\\ 根据题目可列出求解等式：x=1-(2/3)，\\ 因此最终答案为(1/3)。\end{tabular} \\ 
\hline
\end{tabular}}
\end{lrbox}
\scalebox{0.47}{\usebox{\tce}}
\end{table}
\end{CJK*}

\subsubsection{Question \& Answer Corpus}

Q\&A data requires respondents to flexibly organize appropriate responses based on questions, which may enhance the flexibility and adaptability of the model. After training with a corpus of about 400M tokens, we obtain the following observation in Table \ref{tab:dtc}: 1) The two models have similar trends in S/N and J/P but slightly different in F-value. Bloom increases from 11 to 12, but Llama decreases from 15 to 11. 
Results have shown that the Q\&A corpus can enhance the model's adaptability and flexibility (indicated by P-value).

\subsubsection{Examination Corpus}

To improve the thinking ability of the model, we attempt to train the model using an examination corpus (about 50M tokens),  mainly composed of APE210k \cite{zhao2020ape210k} . The results in Table \ref{tab:dtc} show that this corpus has the potential to enhance the T-value significantly. The increase in Bloom's T-value from 13 to 15 and Llama from 9 to 10 demonstrates the effectiveness of the examination corpus in enhancing thinking dichotomy. This enhancement further strengthens Bloom's (ISTJ) thinking personality and facilitates the evolution of Llama (INFJ), who has a more emotional personality towards T.

\textbf{Summary Q3:} The type of training corpus can affect the
MBTI type, especially in the dimensions of T/F and J/P.

\subsection{Can MBTI evaluate the model reasonably? (Q4)}
In this section, we discuss the limitations of MBTI and its value as an LLM evaluation metric.

\subsubsection{The MBTI itself is just pseudo-science.}
In fact, as a measurement tool, MBTI has flaws in reliability and validity. In particular, human activities are always affected by different situations and different mental states. This situation will result in the MBTI being a toy tool for humans. However, these facts have not prevented many companies and individuals from using it as a \textbf{rough tool} to help hire employees or choose a career direction. In a sense, as a rough evaluation, it is also partially reasonable.

1) E/I (extraversion/introversion). On this dichotomy, no significant rules are observed, which may indicate that human relationships do not apply to machines. This assertion makes sense because humans do not want AI to be overly social or shy.

2) S/N (sensing/intuition). Similarly, no discernible patterns have been identified in this dichotomy, nor have any plausible hypotheses been proposed to explain its association with LLMs.

3) T/F (thinking/feeling). We posit that the T-value metric is paramount for LLMs, given that GPT-4 and ChatGPT exhibit significantly higher T-values than other models. Furthermore, the utilization of mathematical corpora has been demonstrated to enhance the model's reasoning capabilities, resulting in a corresponding increase in the T-value metric. Thus, we advocate for the T-value metric as a crucial indicator of a model's proficiency.

4) J/P (judging/perceiving). After comparing the values of GPT-4 and ChatGPT, it is evident that GPT-4 has a higher J-value, which accurately reflects the planning abilities present in human personality. As a result, we believe that models with higher J-values possess more significant potential for task decomposition and path planning.

In conclusion, the T/F and J/P dimensions hold significant value and can be considered reliable indicators for evaluating LLMs. These dimensions can provide insights into various aspects, including knowledge distribution after pre-training, the ability to follow instructions, and more.

\subsubsection{What kind of MBTI type is best for LLMs?}

For humans, each MBTI type is a unique personality. However, LLMs with the proper knowledge, reasoning, and planning capabilities may be the best choice for machines serving humans, e.g., INTJ (GPT-4). Of course, in some scenarios (such as role-playing apps), LLMs can change themselves according to user expectations.

\textbf{Summary Q4:} Although MBTI is not a rigorous assessment, it may still serve as a rough indicator for LLMs.

\section{Conclusion}
In this work, we investigate the question: Do LLMs with human-like abilities exhibit human-like personalities? To address this question, we comprehensively examine the MBTI as a preliminary assessment tool for LLMs from various perspectives. After extensive experiments, our observations lead to several key conclusions: 1) LLMs exhibit diverse personalities; 2) LLMs that lack sufficient instruction tuning are resistant to the change of MBTI types but can be influenced by explicit and implicit prompts after tuning; 3) The type of training corpus can impact the MBTI type; 4) While MBTI is not a rigorous assessment, it can serve as a rough indicator.

In the future, we aim to expand our research by integrating additional pre-training datasets. In this regard, we are particularly intrigued by tasks that enhance commonsense comprehension and reasoning abilities, such as math dataset. 

\section*{Limitations}
In regards to personality indicators, there is potential for future research on AGI to utilize a broader range of personality tests for LLMs. However, this topic needs to be explored in this work. Due to resource limitations, our baselines are trained on models around 10B parameters and 400M tokens. More intriguing findings could emerge with the use of larger models and corpus.


\bibliography{emnlp2022}
\bibliographystyle{emnlp2022}


\end{document}